\documentclass[letterpaper, 10 pt, conference]{ieeeconf}  

\IEEEoverridecommandlockouts                              

\usepackage{amsmath} 
\usepackage{amssymb}  
\usepackage{balance}
\usepackage{graphicx}
\usepackage{times}
\usepackage{dsfont}
\usepackage{multicol}
\usepackage{siunitx}
\usepackage[bookmarks=true]{hyperref}
\usepackage{cleveref}
\usepackage{xspace}
\usepackage{xurl}
\usepackage{capt-of}
\usepackage[usenames,dvipsnames,table,xcdraw]{xcolor}

\usepackage{booktabs}  
\usepackage{float} 
\usepackage[flushleft]{threeparttable}
\usepackage{amsmath,amssymb,amsfonts}
\usepackage{arydshln} 

\usepackage{caption}
\usepackage{stfloats}

\usepackage[
    style=ieee,
    natbib=true,
    citestyle=numeric-comp,
    doi=false,
    isbn=false,
    url=false]{biblatex} 

\definecolor{mydarkblue}{rgb}{0,0.08,0.45}
\definecolor{mydarkgreen}{RGB}{0, 139, 69}
\definecolor{mygreen2}{RGB}{0 205 0}
\definecolor{mybrown}{RGB}{139 69 19}

\definecolor{boxblue}{RGB}{79,173,234}

\definecolor{tablepeach}{RGB}{255, 240, 235}

\definecolor{tablepurple}{RGB}{248,235,252}

\definecolor{tableblue}{RGB}{235,241,255}
\hypersetup{
	colorlinks=true,
	linkcolor=blue,
	urlcolor=magenta,
	citecolor=mygreen2,
}

\usepackage{amsfonts}
\usepackage{bm}
\usepackage{amsmath}
\usepackage{mathtools}
\usepackage{multirow}
\usepackage{booktabs}
\usepackage{caption}
\usepackage{wrapfig,lipsum,booktabs}
\usepackage{caption}
\usepackage{bbm}
\usepackage{multirow}
\usepackage{pifont}
\usepackage[linesnumbered,ruled,vlined]{algorithm2e}

\SetCommentSty{mycommfont}
\SetAlCapFnt{\small} 
\usepackage{etoc}
\SetAlgorithmName{Algo}{Algo}{}

\renewcommand{\paragraph}[1]{{\vspace{1mm}\noindent \bf #1}.}

\newcommand{\reft}{{\bs{\hat{p}}_{t}}}

\newcommand{\refr}{{\bs{\hat{\theta}}_{t}}}

\newcommand{\reftn}{{\bs{\hat{p}}_{t+1}}}
\newcommand{\refrn}{{\bs{\hat{\theta}}_{t+1}}}

\newcommand{\simav}{{\bs{{\omega}}_{t}}}

\newcommand{\simr}{{\bs{{\theta}}_{t}}}
\newcommand{\simt}{{\bs{{p}}_{t}}}
\newcommand{\goalstate}{{\bs{s}^{\text{g}}_t}}

\newcommand{\selfstate}{{\bs{s}^{\text{p}}_t}}
\newcommand{\state}{{\bs{s}_t}}
\newcommand{\action}{{\bs{a}_t}}
\newcommand{\actionprev}{{\bs{a}_{t-1}}}

\newcommand{\bs}[1]{\boldsymbol{#1}}
\newcommand{\xmark}{\ding{55}}%

\newcommand{\method}{HOVER\xspace}

\title{\LARGE \bf
\method: Versatile Neural Whole-Body Controller for Humanoid Robots
}

\author{Tairan He$^{*1,2}$ \quad Wenli Xiao$^{*1,2}$ \quad Toru Lin$^{1,3}$ \quad Zhengyi Luo$^{1,2}$ \quad Zhenjia Xu$^{1}$ \quad Zhenyu Jiang$^{1,4}$ \\ Jan Kautz$^{1}$ \quad  Changliu Liu$^{2}$ \quad Guanya Shi$^{2}$ \quad Xiaolong Wang$^{1,5}$ \quad Linxi ``Jim" Fan$^{\dag 1}$ \quad Yuke Zhu$^{\dag 1,4}$ 
\thanks{*Equal Contributions, $\dag$GEAR Team Leads}
\thanks{$^{1}$NVIDIA, $^{2}$CMU, $^{3}$UC Berkeley, $^{4}$UT Austin, $^{5}$UC San Diego}
\thanks{Paper website: \href{https://hover-versatile-humanoid.github.io}{https://hover-versatile-humanoid.github.io}}
}

\begin{document}

\maketitle
\thispagestyle{empty}
\pagestyle{empty}

\begin{abstract}

Humanoid whole-body control requires adapting to diverse tasks such as navigation, loco-manipulation, and tabletop manipulation, each demanding a different mode of control. For example, navigation relies on root velocity or position tracking, while tabletop manipulation prioritizes upper-body joint angle tracking. Existing approaches typically train individual policies tailored to a specific command space, limiting their transferability across modes. We present the key insight that full-body kinematic motion imitation can serve as a common abstraction for all these tasks and provide general-purpose motor skills for learning multiple modes of whole-body control. Building on this, we propose \method (Humanoid Versatile Controller), a multi-mode policy distillation framework that consolidates diverse control modes into a unified policy. \method enables seamless transitions between control modes while preserving the distinct advantages of each, offering a robust and scalable solution for humanoid control across a wide range of modes. By eliminating the need for policy retraining for each control mode, our approach improves efficiency and flexibility for future humanoid applications.


\end{abstract}

\section{INTRODUCTION}

Humanoid is a versatile form factor that supports a wide variety of robotic tasks and applications, including bimanual manipulation \cite{cheng2024open,lin2024learning,li2024okami}, bipedal locomotion \cite{radosavovic2024real,li2024reinforcement,zhuang2024humanoid,liao2024berkeley}, and agile whole-body control \cite{bdparkour,he2024omnih2o,he2024learning,xue2024full,cheng2024expressive,fu2024humanplus,zhang2024wococo}.
While showing impressive results, each of these efforts uses a different formulation for whole-body control based on the need for their specific task and scenario. Some use root velocity tracking~\cite{li2024reinforcement,zhuang2024humanoid} to support locomotion, some choose joint angle tracking \cite{cheng2024expressive,fu2024humanplus} to enable expressive movements, and others use kinematic tracking of selected body keypoints \cite{he2024learning,he2024omnih2o} to support teleoperation. Although these approaches are similar in terms of the end goal of motion tracking, they require task-specific controller interface and rewards design. This not only makes the development process repetitive and time-consuming, but also limits the versatility of the resultant whole-body controller. For instance, a robot performing bipedal locomotion on uneven terrain using root velocity tracking~\cite{li2024reinforcement,zhuang2024humanoid} would struggle to seamlessly switch to a task requiring precise bimanual manipulation, where joint angle or end-effector tracking~\cite{cheng2024expressive,fu2024humanplus,lin2024learning} might be necessary. 
These task-specific dependencies limit versatility, as each controller is restricted to a single mode of control.
In addition to motion tracking, many pretrained manipulation policies~\cite{padalkar2023open,kim2024openvla} require operating in different configuration spaces, such as joint angles and end-effector positions. This variability highlights the need for a unified low-level humanoid controller capable of adapting to diverse control mode configurations.

\begin{figure}[t]
    \centering
    \includegraphics[width=1.0\columnwidth]{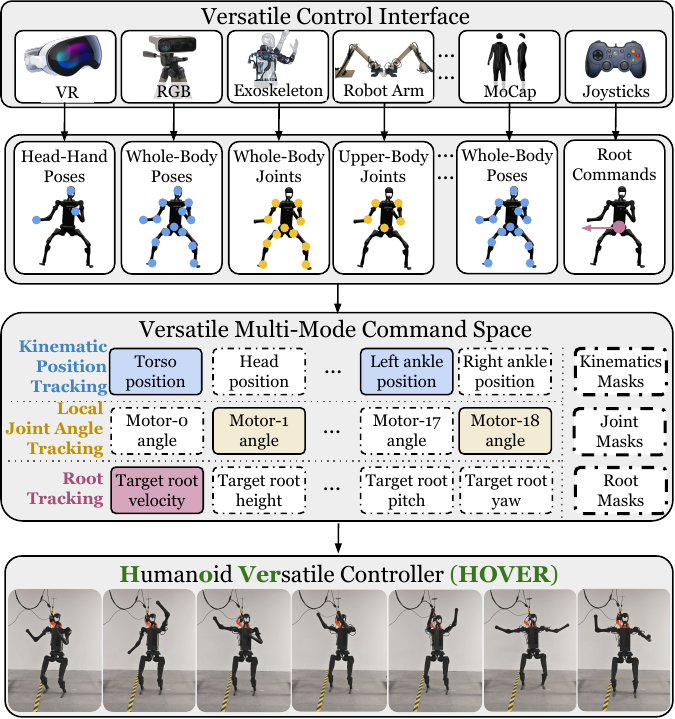}
    \caption{\method enables versatile humanoid control with a unified multi-mode command space. The versatile multi-mode command space supports \textcolor{NavyBlue}{\textbf{kinematic position tracking (blue)}}, \textcolor{Peach}{\textbf{local joint angle tracking (yellow)}}, and \textcolor{Purple}{\textbf{root tracking (purple)}}. Highlighted boxes indicate active commands being tracked, while the masks (dashed boxes on the right) allow selective activation of different command spaces to accommodate various tasks. 
    }
    \label{fig:firstpage}
    \vspace{-15pt}
\end{figure}


Since all these modes are applied to a shared hardware platform, a natural question arises: \textit{Can we create a unified controller that supports all control modes, combining the strengths of each?} This is a non-trivial challenge, as each mode operates within a distinct command space, making direct integration impractical. However, despite differences in control interfaces, the underlying motion objectives often align: stable, human-like motion for humanoid control.

\begin{figure}[tbp]
    \centering
    \includegraphics[width=\columnwidth]{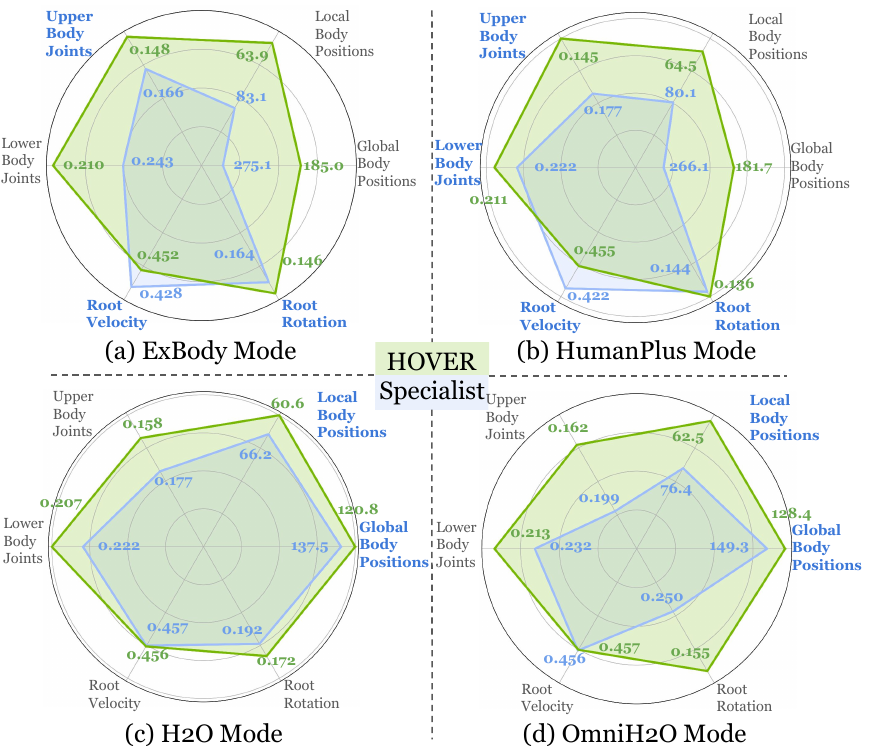} 
    \caption{Comparison between \textcolor{NavyBlue}{\textbf{prior work specialists (blue)}} and our \textcolor{OliveGreen}{\textbf{generalist policy (green)}} under corresponding modes. The metrics used are: upper/lower joint error (rad), global/local body position error (mm), root velocity error (m/s), and root rotation error (rad). These metrics evaluate how accurately each policy tracks reference motions and joint configurations across different control modes. The modes being tracked (activated) by each mode are colored \textcolor{NavyBlue}{\textbf{blue}}.
    }
    \label{fig:radar_specialist}
    \vspace{-17pt}
\end{figure}

To this end, we present \method, a unified neural controller for humanoid whole-body control that supports diverse control modes as shown in \Cref{fig:firstpage}, including over 15 useful modes for real-world applications to a 19-DOF humanoid robot. 
This versatile command space covers most modes used in prior works~\cite{cheng2024expressive,he2024learning,he2024omnih2o,fu2024humanplus}. 
To ensure a robust foundation of motor skills that generalize well across tasks, we train an oracle motion imitator to mimic large-scale human motion data from MoCap~\cite{mahmood2019amass}, covering a wide variety of movements and control objectives. This design choice leverages the inherent adaptability and natural efficiency of human movements, providing the policy with rich motor priors that can be reused across multiple control modes. By grounding the training process in human-like motion, the policy gains a deeper understanding of balance, coordination, and motion control, which are crucial for effective whole-body humanoid behavior.
Through a policy distillation process, we transfer these motor skills from the oracle policy into a single ``generalist policy'' capable of handling multiple control modes. The resulting multi-mode policy not only supports diverse control modes but also \textit{outperforms} policies trained individually for each mode as shown in \Cref{fig:radar_specialist}. We hypothesize that this is due to the policy leveraging shared physical knowledge across modes, such as maintaining balance, human-like motion, and precise limb control. These shared skills enhance generalization, leading to better performance across all modes. In contrast, single-mode policies often overfit to specific reward structures and training environments, limiting their adaptability. Our multi-mode generalist policy also enables seamless transitions between modes, making it both robust and versatile.

To summarize, our contributions are threefold: 1) we present \method, a unified neural controller for humanoid whole-body control supporting multiple control modes; 2) we show that, through policy distillation, \method effectively shares motor skills across modes and outperforms individually trained policies; and 3) experiments in both simulation and on a real humanoid robot demonstrate that \method achieves seamless transitions between modes and delivers superior multi-mode control compared to other baselines.

\begin{figure*}[tbp]
    \centering
    \includegraphics[width=1.0\textwidth]{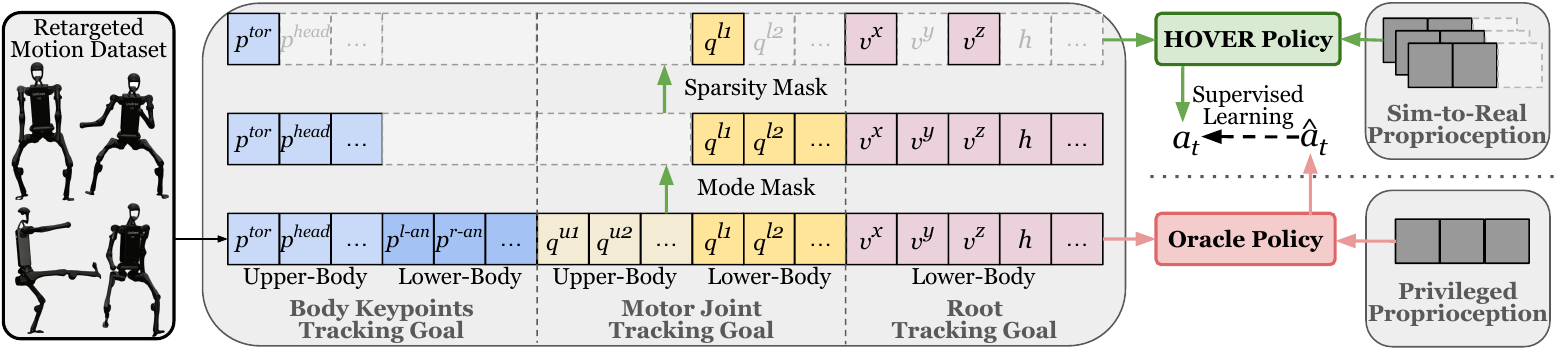}
    \caption{\textbf{Overview of \method distillation process.} The \textcolor{OliveGreen}{\textbf{\method policy}} is distilled from the \textcolor{OrangeRed}{\textbf{Oracle policy}} through proprioception and command masking. The task commands for the student are determined via \textit{mode-specific} and \textit{sparsity-based} masks, applied to both upper and lower body motions independently. These masks generate diverse task command modes, refining the student's inputs. The distillation employs DAgger to align the student’s actions with those of the oracle, optimizing through supervised learning on the oracle’s actions.
    }
    \label{fig:distill_pipeline}
    \vspace{-13pt}
\end{figure*}

\section{METHOD}

\subsection{Goal-Conditioned RL for Humanoid Control} 
We formulate our problem as a goal-conditioned reinforcement learning (RL) task, where the policy $\pi$ is trained to track real-time human motion. 
The state $\state$ comprises both the agent’s proprioception $\selfstate$ and the target goal state $\goalstate$. The goal state $\goalstate$ provides a unified representation of the target motion goal, which we will describe in detail in \Cref{SEC:CommandSpaceDesign}. Using the agent’s proprioception $\selfstate$ and the goal state $\goalstate$, we define the reward $r_t = \mathcal{R}\left(\selfstate, \goalstate\right)$ for policy optimization. The action $\action \in \mathbb{R}^{19}$ represents the target joint positions, which are fed into the PD controller to actuate the robot's degrees of freedom. We employ the proximal policy optimization (PPO) algorithm~\cite{schulman2017proximal} to maximize the cumulative discounted reward $\mathbb{E}\left[\sum_{t=1}^T \gamma^{t-1} r_t\right]$. This setup is framed as a command-tracking task, where the humanoid learns to follow the target commands at each timestep.

\subsection{Command Space Design for Humanoid Control}
\label{SEC:CommandSpaceDesign}
In legged locomotion, root velocity~\cite{margolis2024rapid} or position tracking~\cite{he2024agile} is a commonly employed command space. However, focusing solely on root tracking imposes limitations on the full capabilities of humanoid robots, especially for whole-body loco-manipulation tasks. We observe that while prior works~\cite{cheng2024expressive,he2024learning,he2024omnih2o,fu2024humanplus} have introduced control modes with varying advantages and disadvantages, each is typically tailored to specific subsets of tasks, thus lacking the flexibility required for general-purpose humanoid control. In contrast, our goal is to design a comprehensive control framework that accommodates a wide range of scenarios and is adaptable to various humanoid tasks.
To achieve this, the command space must be constructed to satisfy the following key criteria: 
\begin{itemize}
\item \textbf{Generality}: The command space should encompass most existing configurations, allowing a general-purpose controller to replace task-specific controllers without sacrificing performance or versatility. And the space should be sufficiently expressive to interface with real-world control devices, including joysticks, keyboards, motion capture systems, exoskeletons, and virtual reality (VR) headsets as shown in \Cref{fig:firstpage}.
\item \textbf{Atomicity}: The command space should be composed of independent dimensions, enabling arbitrary combinations of control options to support various modes.   
\end{itemize}

Based on these criteria, we define a unified command space for humanoid whole-body control. This space consists of two primary control regions—upper-body and lower-body control—and incorporates three distinct control modes: 
\begin{itemize} \item \textcolor{NavyBlue}{\textbf{Kinematic Position Tracking}}: target 3D positions of key rigid body points  on the robot. 
\item \textcolor{Peach}{\textbf{Local Joint Angle Tracking}}: target joint angles for each robot motor. 
\item \textcolor{Purple}{\textbf{Root Tracking}}: target root velocity, height, and orientation, specified by roll, pitch, and yaw angles. \end{itemize}

In our framework, as shown in \Cref{fig:firstpage}, a one-hot masking vector is introduced to specify which components of the command space are activated for tracking. Recent work on learning-based humanoid whole-body control~\cite{cheng2024expressive,he2024learning,he2024omnih2o,fu2024humanplus}, as shown in \Cref{tab:existing_command_space} can be viewed as subsets of our unified command space, each representing specific configurations.

\begin{table*}[tbh]
\center
\caption{Command space for priors works on whole-body humanoid control. \method covers all the command designs designed by prior works with a unified command space and supports multi-mode control by tracking arbitrary subsets of the command elements.}
\centering
\begin{threeparttable}
\begin{tabular}{c|cccccc}
\toprule
\multirow{2}{*}{Controller}  &
\multirow{2}{*}{Multi-Mode Control} &
\multicolumn{2}{c}{Upper-Body Command} &
\multicolumn{3}{c}{Lower-Body Command} \\
\cmidrule(lr){3-4} \cmidrule(lr){5-7} 
& &  \textcolor{NavyBlue}{Kinematic Position}  & \textcolor{Peach}{Joint Angle} & \textcolor{NavyBlue}{Kinematic Position}  & \textcolor{Peach}{Joint Angle} & \textcolor{Purple}{Root} \\
\hline
ExBody~\cite{cheng2024expressive} & \textcolor{BrickRed}{\xmark} & \textcolor{BrickRed}{\xmark} & \textcolor{OliveGreen}{\checkmark} &  \textcolor{BrickRed}{\xmark}  & \textcolor{BrickRed}{\xmark} & \textcolor{OliveGreen}{\checkmark}\\
H2O~\cite{he2024learning} & \textcolor{BrickRed}{\xmark} & \textcolor{OliveGreen}{\checkmark} & \textcolor{BrickRed}{\xmark} & \textcolor{OliveGreen}{\checkmark} & \textcolor{BrickRed}{\xmark} & \textcolor{BrickRed}{\xmark}\\
OmniH2O~\cite{he2024omnih2o} & \textcolor{BrickRed}{\xmark} &  \textcolor{OliveGreen}{\checkmark} & \textcolor{BrickRed}{\xmark} &  \textcolor{BrickRed}{\xmark}  & \textcolor{BrickRed}{\xmark} & \textcolor{BrickRed}{\xmark}\\
HumanPlus~\cite{fu2024humanplus} & \textcolor{BrickRed}{\xmark} & \textcolor{BrickRed}{\xmark} & \textcolor{OliveGreen}{\checkmark} &  \textcolor{BrickRed}{\xmark}  & \textcolor{OliveGreen}{\checkmark} & \textcolor{OliveGreen}{\checkmark}\\ 
MHC~\cite{dugar2024learning} & \textcolor{OliveGreen}{\checkmark} & \textcolor{BrickRed}{\xmark} & \textcolor{OliveGreen}{\checkmark} &  \textcolor{BrickRed}{\xmark}  & \textcolor{OliveGreen}{\checkmark} & \textcolor{OliveGreen}{\checkmark}\\ 
\hdashline 
HOVER (ours) & \textcolor{OliveGreen}{\checkmark} & \textcolor{OliveGreen}{\checkmark} & \textcolor{OliveGreen}{\checkmark} &  \textcolor{OliveGreen}{\checkmark}  & \textcolor{OliveGreen}{\checkmark} & \textcolor{OliveGreen}{\checkmark}\\

\bottomrule
\end{tabular}
\end{threeparttable}
\label{tab:existing_command_space}
\vspace{-10pt}
\end{table*}


\subsection{Motion Retargeting}
Recent works have shown the advantage of learning robust whole-body control for humanoid robots from large motion datasets~\cite{cheng2024expressive,he2024learning,he2024omnih2o,fu2024humanplus}.
The retargeting procedure from human motion dataset~\cite{mahmood2019amass} to humanoid motion dataset has three steps:
\textbf{Step-1}: We first compute the keypoints positions of the humanoid using forward kinematics, mapping its joint configurations to workspace coordinates.
\textbf{Step-2}: Next, we fit the SMPL model to match the humanoid’s kinematics by optimizing the SMPL parameters to align with the computed keypoints from forward kinematics.
\textbf{Step-3}: Finally, the AMASS dataset is retargeted by matching corresponding keypoints between the fitted SMPL model and the humanoid with gradient descent.
We follow the same motion retargeting and ``sim-to-data'' procedure with~\cite{he2024learning} to convert the large-scale human motion dataset~\cite{mahmood2019amass} to dataset $\hat Q$ that only contains feasible motions for humanoids.

\subsection{Oracle Policy Training from Large-Scale Human Motions}
\paragraph{State Space Design}
We train an oracle motion imitator $\pi^\text{oracle}(a_t|s_t^\text{p-oracle}, s_t^\text{g-oracle})$. The proprioception is defined as $s_t^\text{p-oracle} \triangleq [\simt, \simr, \dot p_t, \simav, \actionprev]$, which contains the humanoid rigid-body position $\simt$, orientation $\simr$, linear velocity $\dot p_t$, angular velocity $\simav$, and the previous action $\actionprev$. The goal state is defined as $s_t^\text{g-oracle} \triangleq   [\refrn \ominus \simr, \reftn -  \simt, \bs{\hat{v}}_{t+1} -  \bs{v}_t, \bs{\hat{\omega}}_{t+1} -  \bs{\omega}_t, \bs{\hat\theta_t}, \bs{\hat p_t}]$, which contains the reference pose ($\refr, \reft$) and one-frame difference between the reference and current state for all rigid bodies of the humanoid. We use the same policy network structure with~\cite{he2024omnih2o}, a three-layer MLP with layer dimensions of $[512,256,128]$.

\paragraph{Reward Design and Domain Randomizations}
We formulate the reward $r_t$ as the sum of three components: 1) penalty, 2) regularization, and 3) task rewards, detailed in~\Cref{tab:reward}. 
We follow the same domain randomization in~\cite{he2024omnih2o} to randomize the physical parameters of the simulated environment and humanoids for successful sim-to-real transfer.

\begin{table}[tbp]
\centering
\caption{Reward designs for the oracle policy. 
}
\vspace{-3pt}
\resizebox{0.85\linewidth}{!}{%
\begin{tabular}{ c  c  c  c }
\hline
Term & Weight & Term & Weight \\ \hline
\multicolumn{4}{c}{Penalty} \\ \hline
Torque limits & $-2 $ & DoF position limits & $-1.25e^2$ \\
Termination & $-2.5e^2$ & DoF velocity limits & $-5e^1$ \\ \hline
\multicolumn{4}{c}{Regularization} \\ \hline
DoF acceleration & $-1.1e^{-5}$ & DoF velocity & $-4e^{-3}$ \\
Lower Action rate & $-3$ & Upper Action rate & $-6.25e^{-1}$ \\
Torque & $-1e^{-4}$ & Feet orientation & $-6.25e^{1}$\\
Feet air time & $1e^3$ & Feet contact force & $-7.5e^{-1}$ \\
Stumble & $-1.25e^3$ & Slippage & $-7.5e^1$ \\ 
In the air & $-2e^2$ & Max feet height per step & $-3e^3$ \\ \hline
\multicolumn{4}{c}{Task Reward} \\ \hline
DoF position & $3.2e^1$ & DoF velocity & $1.6e^1$ \\
Body position & $8e^1$ & Body rotation & $2e^1$ \\
Body velocity & $8$ & Body angular velocity & $8$ \\ 
Root velocity & $1e^2$ & Root rotation & $2e^1$ \\ \hline
\end{tabular}
}

\label{tab:reward}
\vspace{-15pt}
\end{table}

\subsection{Multi-Mode Versatile Controller via Distillation}
\paragraph{Proprioception}
For the student policy $\pi^\text{student}(s_t^\text{p-student}, s_t^\text{g-student})$ distilled from the oracle teacher $\pi^\text{oracle}$, the proprioception is defined as $s_t^\text{p-student} \triangleq [q, \dot{q}, \omega^\text{base}, g]_{t-25:t} \cup [a_{t-25:t-1}]$, where $q$ is the joint position, $\dot{q}$ is the joint velocity, $\omega^\text{base}$ is the base angular velocity, $g$ is the gravity vector, and $a$ is the action history. Following~\cite{he2024omnih2o}, we stack these terms over the last 25 steps to represent the student's proprioceptive input.

\paragraph{Command Mask} As illustrated in~\Cref{fig:distill_pipeline}, the task command input for the student policy is defined using mode-based and sparsity-based masking. Specifically, the student's task command input, $s_t^\text{g-student}$, is represented as $s_t^\text{g-student} \triangleq M_\text{sparsity} \odot \left[ M_\text{mode} \odot s_t^\text{g-upper}, M_\text{mode} \odot s_t^\text{g-lower} \right]$. The mode mask, $M_\text{mode}$, selects a specific task command mode for the upper and lower body independently. For instance, the upper body may track kinematic positions, while the lower body focuses on joint angle and root tracking, as shown in \Cref{fig:distill_pipeline}. After the mode-specific masking, the sparsity mask, $M_\text{sparsity}$, is applied. For example, in some scenarios, the upper body may track only the kinematic positions of the hands, while the lower body tracks only the joint angles of the torso. Every bit of the mode and sparsity binary mask is from a Bernoulli distribution $\mathcal{B}(0.5)$. Mode and sparsity masks are randomized at the episode beginning and remain fixed until the episode ends

\paragraph{Policy Distillation} We perform policy distillation using the DAgger framework~\cite{ross2011reduction}. For each episode, we roll out the student policy $\pi^\text{student}(\mathbf{a}_t | s_t^\text{p-student}, s_t^\text{g-student})$ in simulation to obtain trajectories of $(s_t^\text{p-student}, s_t^\text{g-student})$. At each timestep, we also compute the corresponding oracle states $(s_t^\text{p-oracle}, s_t^\text{g-oracle})$. Using these oracle states, we query the oracle teacher policy $\pi^\text{oracle}(\hat{\mathbf{a}}_t | s_t^\text{p-oracle}, s_t^\text{g-oracle})$ to obtain the reference action $\hat{\mathbf{a}}_t$. The student policy $\pi^\text{student}$ is then updated by minimizing the loss function:
$\mathcal{L} = \| \hat{\mathbf{a}}_t - \mathbf{a}_t \|^2_2,$ where $\hat{\mathbf{a}}_t$ is the reference action from the oracle, and $\mathbf{a}_t$ is the action taken by the student policy.

\begin{table*}[tbp]
\caption{Simulation motion imitation evaluation of \method and baselines on dataset $\hat Q$. Metrics that are tracked by different modes are highlighted in corresponding colors. Results that are statistically significant are highlighted in bold across 5 random seeds. 
}
\label{tab:imitation_sim_table}
\centering
\resizebox{\linewidth}{!}{%
\begingroup
\setlength{\tabcolsep}{3pt} 
\renewcommand{\arraystretch}{1.2} 
\begin{tabular}{lcccccccccccc}
\toprule
\multicolumn{1}{c}{} & \multicolumn{5}{c}{\textcolor{NavyBlue}{\textbf{Kinematic Position}}} & \multicolumn{2}{c}{\textcolor{Peach}{\textbf{Joint Angle}}} & \multicolumn{5}{c}{\textcolor{Purple}{\textbf{Root}}} \\ 
\cmidrule(lr){1-1} \cmidrule(lr){2-6} \cmidrule(lr){7-8} \cmidrule(lr){9-13} 

Method & $\text{Survive} \uparrow$ & $E_\text{g-mpjpe} \downarrow$ & $E_\text{mpjpe} \downarrow$ & $\text{E}_{\text{acc}} \downarrow$ & $\text{E}_{\text{vel}} \downarrow$ & $E_\text{upper-j}\downarrow$ & $E_\text{lower-j}\downarrow$ & $E_\text{root-vel}\downarrow$ & $E_\text{root-h}\downarrow$ & $E_\text{root-r}\downarrow$ & $E_\text{root-p}\downarrow$ & $E_\text{root-y}\downarrow$ \\ 
\cmidrule(lr){1-1} \cmidrule(lr){2-6} \cmidrule(lr){7-8} \cmidrule(lr){9-13}

Oracle policy & {\textcolor{black}{99.3\%\(\tiny{\pm \text{0.203}}\)}} & {\textcolor{black}{119\(\tiny{\pm \text{0.442}}\)}} & {\textcolor{black}{59.4\(\tiny{\pm \text{0.234}}\)}} & {\textcolor{black}{2.63\(\tiny{\pm \text{0.008}}\)}} & {\textcolor{black}{5.43\(\tiny{\pm \text{0.024}}\)}} & {\textcolor{black}{0.153\(\tiny{\pm \text{0.001}}\)}} & {\textcolor{black}{0.206\(\tiny{\pm \text{0.001}}\)}} & {\textcolor{black}{0.456\(\tiny{\pm \text{0.002}}\)}} & {\textcolor{black}{0.066\(\tiny{\pm \text{0.001}}\)}} & {\textcolor{black}{0.065\(\tiny{\pm \text{0.001}}\)}} & {\textcolor{black}{0.083\(\tiny{\pm \text{0.001}}\)}} & {\textcolor{black}{0.282\(\tiny{\pm \text{0.002}}\)}}\\
\cmidrule(lr){1-1} \cmidrule(lr){2-6} \cmidrule(lr){7-8} \cmidrule(lr){9-13} 
\multicolumn{13}{l}{\textbf{ExBody Mode} - Upper: \textcolor{Peach}{joint angle tracking}, Lower: \textcolor{Purple}{root tracking}} \\
\cmidrule(lr){1-1} \cmidrule(lr){2-6} \cmidrule(lr){7-8} \cmidrule(lr){9-13} 
ExBody (Specialist) & {\textcolor{black}{99.1\%\(\tiny{\pm \text{0.212}}\)}} & {\textcolor{black}{275\(\tiny{\pm \text{1.650}}\)}} & {\textcolor{black}{83.1\(\tiny{\pm \text{0.499}}\)}} & \textbf{\textcolor{black}{2.63\(\tiny{\pm \text{0.007}}\)}} & {\textcolor{black}{6.31\(\tiny{\pm \text{0.034}}\)}} & \cellcolor{tablepeach}{\textcolor{black}{0.166\(\tiny{\pm \text{0.002}}\)}} & {\textcolor{black}{0.243\(\tiny{\pm \text{0.003}}\)}} & \cellcolor{tablepurple}\textbf{\textcolor{black}{0.428\(\tiny{\pm \text{0.007}}\)}} & \cellcolor{tablepurple}{\textcolor{black}{0.074\(\tiny{\pm \text{0.001}}\)}} & \cellcolor{tablepurple}{\textcolor{black}{0.070\(\tiny{\pm \text{0.001}}\)}} & \cellcolor{tablepurple}{\textcolor{black}{0.147\(\tiny{\pm \text{0.002}}\)}} & \cellcolor{tablepurple}{\textcolor{black}{0.276\(\tiny{\pm \text{0.003}}\)}} \\
\method (Ours) & {\textcolor{black}{99.1\%\(\tiny{\pm \text{0.220}}\)}} & \textbf{\textcolor{black}{185\(\tiny{\pm \text{1.110}}\)}} & \textbf{\textcolor{black}{63.9\(\tiny{\pm \text{0.384}}\)}} & {\textcolor{black}{3.01\(\tiny{\pm \text{0.009}}\)}} & \textbf{\textcolor{black}{6.06\(\tiny{\pm \text{0.033}}\)}} & \cellcolor{tablepeach}\textbf{\textcolor{black}{0.148\(\tiny{\pm \text{0.002}}\)}} & \textbf{\textcolor{black}{0.210\(\tiny{\pm \text{0.004}}\)}} &\cellcolor{tablepurple}{\textcolor{black}{0.452\(\tiny{\pm \text{0.006}}\)}} & \cellcolor{tablepurple}\textbf{\textcolor{black}{0.063\(\tiny{\pm \text{0.001}}\)}} & \cellcolor{tablepurple}{\textcolor{black}{0.068\(\tiny{\pm \text{0.001}}\)}} & \cellcolor{tablepurple}\textbf{\textcolor{black}{0.091\(\tiny{\pm \text{0.001}}\)}} & \cellcolor{tablepurple}{\textcolor{black}{0.279\(\tiny{\pm \text{0.002}}\)}} \\

\cmidrule(lr){1-1} \cmidrule(lr){2-6} \cmidrule(lr){7-8} \cmidrule(lr){9-13} 

\multicolumn{13}{l}{\textbf{HumanPlus Mode}- Upper: \textcolor{Peach}{joint angle tracking}, Lower: \textcolor{Peach}{joint angle tracking}, \textcolor{Purple}{root tracking}} \\
\cmidrule(lr){1-1} \cmidrule(lr){2-6} \cmidrule(lr){7-8} \cmidrule(lr){9-13} 
HumanPlus (Specialist) & {\textcolor{black}{98.4\%\(\tiny{\pm \text{0.259}}\)}} & {\textcolor{black}{266\(\tiny{\pm \text{1.597}}\)}} & {\textcolor{black}{80.1\(\tiny{\pm \text{0.481}}\)}} & \textbf{\textcolor{black}{2.53\(\tiny{\pm \text{0.007}}\)}} & {\textcolor{black}{6.16\(\tiny{\pm \text{0.033}}\)}} & \cellcolor{tablepeach}{\textcolor{black}{0.177\(\tiny{\pm \text{0.002}}\)}} & \cellcolor{tablepeach}{\textcolor{black}{0.222\(\tiny{\pm \text{0.002}}\)}} &  \cellcolor{tablepurple}\textbf{\textcolor{black}{0.422\(\tiny{\pm \text{0.006}}\)}} &  \cellcolor{tablepurple}\textbf{\textcolor{black}{0.061\(\tiny{\pm \text{0.001}}\)}} &  \cellcolor{tablepurple}{\textcolor{black}{0.080\(\tiny{\pm \text{0.001}}\)}} &  \cellcolor{tablepurple}{\textcolor{black}{0.124\(\tiny{\pm \text{0.001}}\)}} & \cellcolor{tablepurple} {\textcolor{black}{0.228\(\tiny{\pm \text{0.002}}\)}} \\
\method (Ours) & \textcolor{black}{98.9\%\(\tiny{\pm \text{0.285}}\)} & \textbf{\textcolor{black}{182\(\tiny{\pm \text{1.093}}\)}} & \textbf{\textcolor{black}{64.5\(\tiny{\pm \text{0.387}}\)}} & {\textcolor{black}{2.85\(\tiny{\pm \text{0.008}}\)}} & \textbf{\textcolor{black}{5.91\(\tiny{\pm \text{0.032}}\)}} & \cellcolor{tablepeach}\textbf{\textcolor{black}{0.145\(\tiny{\pm \text{0.001}}\)}} & \cellcolor{tablepeach}\textbf{\textcolor{black}{0.211\(\tiny{\pm \text{0.002}}\)}} &  \cellcolor{tablepurple}{\textcolor{black}{0.455\(\tiny{\pm \text{0.007}}\)}} &  \cellcolor{tablepurple}{\textcolor{black}{0.067\(\tiny{\pm \text{0.001}}\)}} &  \cellcolor{tablepurple}\textbf{\textcolor{black}{0.069\(\tiny{\pm \text{0.001}}\)}} &  \cellcolor{tablepurple}\textbf{\textcolor{black}{0.104\(\tiny{\pm \text{0.001}}\)}} &  \cellcolor{tablepurple}{\textcolor{black}{0.237\(\tiny{\pm \text{0.002}}\)}} \\
\cmidrule(lr){1-1} \cmidrule(lr){2-6} \cmidrule(lr){7-8} \cmidrule(lr){9-13} 
\multicolumn{13}{l}{\textbf{H2O Mode} - Upper: \textcolor{NavyBlue}{kinematic position tracking} (left/right hand, left/right shoulder, left/right elbow), Lower: \textcolor{NavyBlue}{kinematic position tracking} (left/right ankle)} \\
\cmidrule(lr){1-1} \cmidrule(lr){2-6} \cmidrule(lr){7-8} \cmidrule(lr){9-13} 

H2O (Specialist) & \cellcolor{tableblue}\textcolor{black}{99.2\%\(\tiny{\pm \text{0.233}}\)} & \cellcolor{tableblue}{\textcolor{black}{137\(\tiny{\pm \text{0.827}}\)}} & \cellcolor{tableblue}{\textcolor{black}{66.3\(\tiny{\pm \text{0.398}}\)}} & \cellcolor{tableblue}\textbf{\textcolor{black}{2.63\(\tiny{\pm \text{0.008}}\)}} & \cellcolor{tableblue}{\textcolor{black}{5.75\(\tiny{\pm \text{0.028}}\)}} & {\textcolor{black}{0.177\(\tiny{\pm \text{0.003}}\)}} & {\textcolor{black}{0.221\(\tiny{\pm \text{0.002}}\)}} & {\textcolor{black}{0.457\(\tiny{\pm \text{0.004}}\)}} & {\textcolor{black}{0.078\(\tiny{\pm \text{0.001}}\)}} & {\textcolor{black}{0.067\(\tiny{\pm \text{0.001}}\)}} & {\textcolor{black}{0.095\(\tiny{\pm \text{0.001}}\)}} & {\textcolor{black}{0.415\(\tiny{\pm \text{0.003}}\)}} \\
\method (Ours) &\cellcolor{tableblue} {\textcolor{black}{98.9\%\(\tiny{\pm \text{0.276}}\)}} & \cellcolor{tableblue}\textbf{\textcolor{black}{121\(\tiny{\pm \text{0.726}}\)}} & \cellcolor{tableblue}\textbf{\textcolor{black}{60.6\(\tiny{\pm \text{0.361}}\)}} & \cellcolor{tableblue}{\textcolor{black}{2.73\(\tiny{\pm \text{0.008}}\)}} & \cellcolor{tableblue}\textbf{\textcolor{black}{5.49\(\tiny{\pm \text{0.028}}\)}} & \textbf{\textcolor{black}{0.158\(\tiny{\pm \text{0.002}}\)}} & \textbf{\textcolor{black}{0.207\(\tiny{\pm \text{0.002}}\)}} & {\textcolor{black}{0.456\(\tiny{\pm \text{0.005}}\)}} & \textbf{\textcolor{black}{0.065\(\tiny{\pm \text{0.001}}\)}} & {\textcolor{black}{0.067\(\tiny{\pm \text{0.001}}\)}} & \textbf{\textcolor{black}{0.086\(\tiny{\pm \text{0.001}}\)}} & \textbf{\textcolor{black}{0.365\(\tiny{\pm \text{0.003}}\)}} \\
\cmidrule(lr){1-1} \cmidrule(lr){2-6} \cmidrule(lr){7-8} \cmidrule(lr){9-13} 

\multicolumn{13}{l}{\textbf{OmniH2O Mode} - Upper: \textcolor{NavyBlue}{kinematic position tracking} (head, left/right hand), Lower: N/A} \\
\cmidrule(lr){1-1} \cmidrule(lr){2-6} \cmidrule(lr){7-8} \cmidrule(lr){9-13} 

OmniH2O (Specialist) & \cellcolor{tableblue}{\textcolor{black}{99.0\%\(\tiny{\pm \text{0.301}}\)}} & \cellcolor{tableblue}{\textcolor{black}{149\(\tiny{\pm \text{0.897}}\)}} & \cellcolor{tableblue}{\textcolor{black}{76.4\(\tiny{\pm \text{0.459}}\)}} & \cellcolor{tableblue}{\textcolor{black}{2.69\(\tiny{\pm \text{0.007}}\)}} & \cellcolor{tableblue}{\textcolor{black}{6.18\(\tiny{\pm \text{0.037}}\)}} & {\textcolor{black}{0.199\(\tiny{\pm \text{0.002}}\)}} & {\textcolor{black}{0.232\(\tiny{\pm \text{0.003}}\)}} & {\textcolor{black}{0.456\(\tiny{\pm \text{0.002}}\)}} & {\textcolor{black}{0.071\(\tiny{\pm \text{0.001}}\)}} & {\textcolor{black}{0.070\(\tiny{\pm \text{0.001}}\)}} & {\textcolor{black}{0.125\(\tiny{\pm \text{0.002}}\)}} & \textbf{\textcolor{black}{0.306\(\tiny{\pm \text{0.002}}\)}} \\
\method (Ours) & \cellcolor{tableblue}{\textcolor{black}{99.0\%\(\tiny{\pm \text{0.297}}\)}} & \cellcolor{tableblue}\textbf{\textcolor{black}{128\(\tiny{\pm \text{0.768}}\)}} & \cellcolor{tableblue}\textbf{\textcolor{black}{62.5\(\tiny{\pm \text{0.368}}\)}} & \cellcolor{tableblue}{\textcolor{black}{2.69\(\tiny{\pm \text{0.008}}\)}} & \cellcolor{tableblue}\textbf{\textcolor{black}{5.65\(\tiny{\pm \text{0.032}}\)}} & \textbf{\textcolor{black}{0.162\(\tiny{\pm \text{0.002}}\)}} & \textbf{\textcolor{black}{0.213\(\tiny{\pm \text{0.002}}\)}} & {\textcolor{black}{0.457\(\tiny{\pm \text{0.004}}\)}} & \textbf{\textcolor{black}{0.065\(\tiny{\pm \text{0.001}}\)}} & {\textcolor{black}{0.068\(\tiny{\pm \text{0.001}}\)}} & \textbf{\textcolor{black}{0.089\(\tiny{\pm \text{0.001}}\)}} & {\textcolor{black}{0.310\(\tiny{\pm \text{0.002}}\)}} \\

\cmidrule(lr){1-1} \cmidrule(lr){2-6} \cmidrule(lr){7-8} \cmidrule(lr){9-13} 

\bottomrule 
\end{tabular}
\endgroup
}
\vspace{-16pt}
\end{table*}

\section{EXPERIMENT}

In this section, we present extensive experimental results in both IsaacGym~\cite{makoviychuk2021isaac} and the real-world Unitree H1~\cite{unitreeh1} robot to address the following questions:
\begin{itemize}
\item \textbf{Q1}: Can \method as a generalist policy outperform policies trained for a specific command configuration?
\item \textbf{Q2}: Can \method outperform other methods of training a multi-mode humanoid controller?
\item \textbf{Q3}: Can \method transfer to real-world hardware and execute versatile multi-mode control?
\end{itemize}

\paragraph{Experiment Setup} 
To answer these questions, we evaluate \method on motion tracking in both simulation (\Cref{sec:exp-q1} and~\Cref{sec:exp-q2}) and real-world settings (\Cref{sec:exp-q3}). In simulation, we evaluate using the retargeted AMASS dataset $\hat Q$. In the real world, we test 20 standing motion sequences focusing on quantitative tracking and locomotion tasks for qualitative multi-mode control. Our real robot employs a 19-DOF Unitree H1 platform~\cite{unitreeh1} with a total mass of around 51.5kg and a height of around 1.8m.

\paragraph{Baselines} To address \textbf{Q1} and \textbf{Q3}, we compare \method with several specialists. As shown in \Cref{tab:existing_command_space}, ExBody~\cite{cheng2024expressive} focuses on tracking upper body joint angles and root velocity, HumanPlus~\cite{fu2024humanplus} tracks whole-body joints and root velocity, H2O~\cite{he2024learning} tracks the kinematic positions of eight keypoints (shoulders, elbows, hands, ankles), and OmniH2O~\cite{he2024omnih2o} tracks the kinematic positions of the head and both hands. We also compare other useful tracking modes  (e.g., left-hand mode, right-hand mode, two-hand mode, head mode). For each control mode, we provide only the relevant observation input to the controller and train the specialist baseline with RL. For instance, in left-hand-only mode, only reference motion of the left hand is provided. To address \textbf{Q2}, we compare with another multi-mode RL policy, which follows the same masking process on the goal commands, but trains the baseline with RL objective from scratch. During the multi-mode RL baseline training, mode and sparsity are randomized at the beginning of each episode and remain fixed until the episode ends, which is the same as the randomized masking process during distillation.

\paragraph{Metrics}
We report survival rate, where the episode terminates if the humanoid hits the ground, not by feet. We calculate tracking error in terms of kinematic pose, joint angles, and root twist and rotations. The mean values of the metrics are computed across all motion sequences from dataset $\hat Q$. We evaluate policy’s ability to imitate the reference motion by compare the tracking error of the global body position $E_{g-\text{mpjpe}}$ (mm), the root-relative mean per-joint (MPJPE) $E_{\text{mpjpe}}$ (mm), joint tracking error $E_j$ (rad), root velocity $E_\text{root-vel}$ (m/s), and root orientation tracking error $E_\text{root-rpy}$ (rad).
To show physical realism, we report average joint acceleration  $E_{\text{acc}}$ $\text{(mm/frame}^2)$, and velocity $E_{\text{vel}}$ (mm/frame) error. 
To better show the correspondence between control modes and metrics, we highlight the metrics that are actively tracked by each mode with corresponding colors in \Cref{tab:imitation_sim_table}, \Cref{tab:comparison_metrics} and \Cref{tab:realworld_metrics}. 
For instance, in~\Cref{tab:imitation_sim_table}, upper joint and root metrics are colored with corresponding mode for ExBody mode.

\subsection{Comparison with Specialists}
\label{sec:exp-q1}
\paragraph{Comparison with Specialists of Prior Work's Control Mode} 
To address \textbf{Q1} (\textit{Can \method as a generalist policy outperform policies trained for a specific command configuration?}),
we compare the performance of the same \method policy across different control modes against the corresponding specialist policies. For example, the performance of \method under ExBody mode is evaluated with a fixed mask to match ExBody mode across the entire dataset $\hat Q$.
As shown in ~\Cref{tab:imitation_sim_table} and \Cref{fig:radar_specialist}, \method consistently demonstrates superior generalization. In every command mode, \method outperforms prior work specialist controllers in at least 7 out of the 12 metrics, as highlighted by the bold values in~\Cref{tab:imitation_sim_table}. 
This consistent advantage across various control modes underscores the versatility of \method. Furthermore, this means that even when focusing on a single control mode without considering multi-mode versatility, distilling from an oracle policy still surpasses RL-trained specialists.

\paragraph{Comparison with Other Specialists of Other Useful Control Mode} In addition to the aforementioned baselines, we also evaluate four additional modes: left-hand mode, right-hand mode, two-hand mode, and head mode. We train four RL specialists to track these modes individually. The results in~\Cref{tab:comparison_metrics} show that \method consistently outperforms specialists in terms of tracking metrics that are trained for specific command configurations. 

\begin{table}[tbp]
\caption{Comparison between \method and specialists.  We only report tracking metrics that are tracked by this mode.
}
\label{tab:comparison_metrics}
\centering
\resizebox{\linewidth}{!}{%
\begingroup
\setlength{\tabcolsep}{3pt} 
\renewcommand{\arraystretch}{1.2} 
\begin{tabular}{lcccc}
\toprule

Method & $E_\text{g-mpjpe-mode}\downarrow$ & $E_\text{mpjpe-mode}\downarrow$ & $E_\text{acc-mode}\downarrow$ & $E_\text{vel-mode}\downarrow$ \\ 
\cmidrule(lr){1-1} \cmidrule(lr){2-5} 

\multicolumn{5}{l}{\textbf{Left Hand Mode} - Upper: \textcolor{NavyBlue}{kinematic position tracking} (left hand), Lower: N/A} \\
\cmidrule(lr){1-1} \cmidrule(lr){2-5} 
Specialist & \cellcolor{tableblue}{\textcolor{black}{189\(\tiny{\pm \text{1.526}}\)}} & \cellcolor{tableblue}{\textbf{\textcolor{black}{147\(\tiny{\pm \text{1.324}}\)}}} & \cellcolor{tableblue}{\textcolor{black}{5.82\(\tiny{\pm \text{0.029}}\)}} & \cellcolor{tableblue}{\textcolor{black}{11.2\(\tiny{\pm \text{0.089}}\)}} \\
\method (Ours) & \cellcolor{tableblue}\textbf{\textcolor{black}{138\(\tiny{\pm \text{1.025}}\)}} & \cellcolor{tableblue}{\textcolor{black}{151\(\tiny{\pm \text{0.934}}\)}} & \cellcolor{tableblue}{\textbf{\textcolor{black}{5.45\(\tiny{\pm \text{0.031}}\)}}} & \cellcolor{tableblue}\textbf{{\textcolor{black}{10.3\(\tiny{\pm \text{0.104}}\)}}} \\

\cmidrule(lr){1-1} \cmidrule(lr){2-5} 
\multicolumn{5}{l}{\textbf{Right Hand Mode} - Upper: \textcolor{NavyBlue}{kinematic position tracking} (right hand), Lower: N/A} \\
\cmidrule(lr){1-1} \cmidrule(lr){2-5} 
Specialist & \cellcolor{tableblue}{\textcolor{black}{220\(\tiny{\pm \text{1.345}}\)}} & \cellcolor{tableblue}{\textcolor{black}{216\(\tiny{\pm \text{1.451}}\)}} & \cellcolor{tableblue}{\textcolor{black}{6.77\(\tiny{\pm \text{0.051}}\)}} & \cellcolor{tableblue}{\textcolor{black}{12.5\(\tiny{\pm \text{0.152}}\)}} \\
\method (Ours) & \cellcolor{tableblue}\textbf{\textcolor{black}{128\(\tiny{\pm \text{0.774}}\)}} & \cellcolor{tableblue}\textbf{\textcolor{black}{141\(\tiny{\pm \text{0.821}}\)}} & \cellcolor{tableblue}{\textbf{\textcolor{black}{5.83\(\tiny{\pm \text{0.049}}\)}}} & \cellcolor{tableblue}\textbf{{\textcolor{black}{10.8\(\tiny{\pm \text{0.129}}\)}}} \\

\cmidrule(lr){1-1} \cmidrule(lr){2-5} 
\multicolumn{5}{l}{\textbf{2 Hands Mode} - Upper: \textcolor{NavyBlue}{kinematic position tracking} (left-right hands), Lower: N/A} \\
\cmidrule(lr){1-1} \cmidrule(lr){2-5} 
Specialist & \cellcolor{tableblue}{\textcolor{black}{137\(\tiny{\pm \text{0.998}}\)}} & \cellcolor{tableblue}{\textcolor{black}{145\(\tiny{\pm \text{1.010}}\)}} & \cellcolor{tableblue}{\textcolor{black}{5.72\(\tiny{\pm \text{0.037}}\)}} & \cellcolor{tableblue}{\textcolor{black}{11.2\(\tiny{\pm \text{0.004}}\)}} \\
\method (Ours) & \cellcolor{tableblue}\textbf{\textcolor{black}{120\(\tiny{\pm \text{0.901}}\)}} & \cellcolor{tableblue}\textbf{\textcolor{black}{119\(\tiny{\pm \text{0.827}}\)}} & \cellcolor{tableblue}{\textbf{\textcolor{black}{5.60\(\tiny{\pm \text{0.045}}\)}}} & \cellcolor{tableblue}\textbf{{\textcolor{black}{10.1\(\tiny{\pm \text{0.134}}\)}}} \\

\cmidrule(lr){1-1} \cmidrule(lr){2-5} 
\multicolumn{5}{l}{\textbf{Head Mode} - Upper: \textcolor{NavyBlue}{kinematic position tracking} (robot head), Lower: N/A} \\
\cmidrule(lr){1-1} \cmidrule(lr){2-5} 
Specialist  & \cellcolor{tableblue}{\textcolor{black}{186\(\tiny{\pm \text{1.149}}\)}} & \cellcolor{tableblue}{\textcolor{black}{104\(\tiny{\pm \text{0.814}}\)}} & \cellcolor{tableblue}\textbf{{\textcolor{black}{2.22\(\tiny{\pm \text{0.008}}\)}}} & \cellcolor{tableblue}{\textcolor{black}{6.63\(\tiny{\pm \text{0.065}}\)}} \\
\method (Ours) & \cellcolor{tableblue}{\textbf{\textcolor{black}{133\(\tiny{\pm \text{0.849}}\)}}} & \cellcolor{tableblue}{\textbf{\textcolor{black}{80.0\(\tiny{\pm \text{0.711}}\)}}} & \cellcolor{tableblue}{\textcolor{black}{2.31\(\tiny{\pm \text{0.011}}\)}} & \cellcolor{tableblue}\textbf{{\textcolor{black}{6.40\(\tiny{\pm \text{0.029}}\)}}} \\


\bottomrule
\end{tabular}
\endgroup
}
\vspace{-15pt}
\end{table}

\vspace{-2pt}
\subsection{Comparison with Other Generalist Training Methods.}
\label{sec:exp-q2}
\begin{figure}[tbp]
    \centering
    \includegraphics[width=0.95\columnwidth]{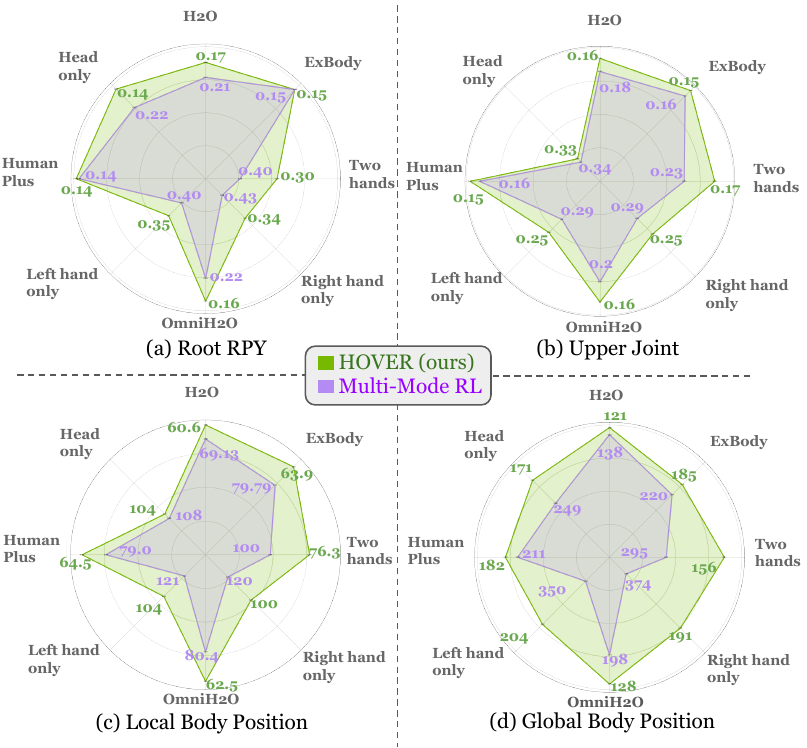} 
    \caption{We assess the tracking accuracy of two multi-mode control policies—\textcolor{OliveGreen}{\textbf{HOVER (green)}} and \textcolor{Purple}{\textbf{Multi-Mode RL (purple)}}—across eight distinct humanoid control modes. The comparison is visualized across four key performance metrics in the radar charts. 
    }
    \label{fig:radar_generalist}
    \vspace{-10pt}
\end{figure}

To address \textbf{Q2} (\textit{does \method outperform other methods of training a multi-mode humanoid controller?}), 
we compare \method with a multi-mode RL baseline that follows the same masking process on the commands but trains with RL objective from scratch.
In~\Cref{fig:radar_generalist}, we assess tracking error across four metrics: root orientation, upper joint angle, and local and global body positions, measured in eight different modes. We scale the tracking error via $\frac{E^\text{max} - E^{(.)}}{E^\text{max}-E^\text{min}}$ for visualization, where larger radar webs indicate better tracking performance. The results show that \method achieves consistently lower tracking error across 32/32 metrics and modes. 
This performance boost underscores the importance of distilling from an oracle policy that tracks full-body kinematics for learning a generalist whole-body controller.

\subsection{Real-World Evaluation}
\label{sec:exp-q3}

To address \textbf{Q3} (\textit{does \method transfer to real-world hardware and execute versatile multi-mode control?}), we conduct quantitative tracking experiments and locomotion tests for qualitative multi-mode control.  

\begin{figure}[tbp]
    \centering
    \includegraphics[width=0.95\columnwidth]{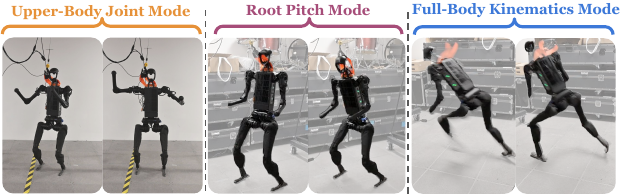} 
    \caption{Real-World Evaluations on different control modes.}
    \label{fig:16-standing-motions}
    \vspace{-17pt}
\end{figure}

\paragraph{Standing Motion Evaluations} We evaluate \method's performance in the real world by tracking 20 different standing motions in $\hat{Q}$. Among these, two motions are visually illustrated in~\Cref{fig:16-standing-motions} (left). The quantitative metrics presented in~\Cref{tab:realworld_metrics} demonstrate \method outperforms specialist policies in 11 out of 12 metrics. 
Moreover, we demonstrate successful tracking of root pitch motion, as shown in the middle of~\Cref{fig:16-standing-motions}, and full-body kinematic tracking, as shown on the right of~\Cref{fig:16-standing-motions}, where the robot is capable of tracking highly dynamic running motions.

\begin{table}[tbp]
\caption{Real-world tracking evaluation on 20 standing motions between prior works specialists and our method. Results that are statistically significant are highlighted in bold across 5 tests.
}
\label{tab:realworld_metrics}
\centering
\resizebox{0.90\linewidth}{!}{%
\begingroup
\setlength{\tabcolsep}{3pt} 
\renewcommand{\arraystretch}{1.2} 
\begin{tabular}{lcccc}
\toprule

Method & $E_\text{g-mpjpe}\downarrow$ & $E_\text{mpjpe}\downarrow$ & $E_\text{upper-j}\downarrow$ & $E_\text{root-rpy}\downarrow$ \\ 
\cmidrule(lr){1-1} \cmidrule(lr){2-5} 

\multicolumn{5}{l}{\textbf{ExBody Mode}} \\
\cmidrule(lr){1-1} \cmidrule(lr){2-5} 
ExBody (Specialist) & {\textcolor{black}{51.3 \(\pm \text{\tiny{0.279}}\)}} & {\textcolor{black}{39.3 \(\pm \text{\tiny{0.214}}\)}} & \cellcolor{tablepeach}{{\textcolor{black}{0.131 \(\pm \text{\tiny{0.001}}\)}}} & \cellcolor{tablepurple}{{\textcolor{black}{0.036 \(\pm \text{\tiny{0.001}}\)}}} \\
\method (Ours) & \textbf{\textcolor{black}{48.9 \(\pm \text{\tiny{0.470}}\)}} & \textbf{\textcolor{black}{36.8 \(\pm \text{\tiny{0.201}}\)}} & \cellcolor{tablepeach}{\textbf{\textcolor{black}{0.126 \(\pm \text{\tiny{0.001}}\)}}} & \cellcolor{tablepurple}{\textbf{\textcolor{black}{0.032 \(\pm \text{\tiny{0.001}}\)}}} \\

\cmidrule(lr){1-1} \cmidrule(lr){2-5} 
\multicolumn{5}{l}{\textbf{HumanPlus Mode}} \\
\cmidrule(lr){1-1} \cmidrule(lr){2-5} 
Specialist & {\textcolor{black}{51.0 \(\pm \text{\tiny{0.275}}\)}} & {\textcolor{black}{36.7 \(\pm \text{\tiny{0.202}}\)}} & \cellcolor{tablepeach}{{\textcolor{black}{0.128 \(\pm \text{\tiny{0.001}}\)}}} &  \cellcolor{tablepurple}{\textbf{\textcolor{black}{0.035 \(\pm \text{\tiny{0.001}}\)}}} \\
\method (Ours) & \textbf{\textcolor{black}{47.4 \(\pm \text{\tiny{0.359}}\)}} & \textbf{\textcolor{black}{35.3 \(\pm \text{\tiny{0.194}}\)}}  & \cellcolor{tablepeach}{\textbf{\textcolor{black}{0.121 \(\pm \text{\tiny{0.001}}\)}}} &  \cellcolor{tablepurple}{{\textcolor{black}{0.038 \(\pm \text{\tiny{0.001}}\)}}} \\

\cmidrule(lr){1-1} \cmidrule(lr){2-5} 
\multicolumn{5}{l}{\textbf{OmniH2O Mode}} \\
\cmidrule(lr){1-1} \cmidrule(lr){2-5} 
Specialist  & \cellcolor{tableblue}{{\textcolor{black}{51.2 \(\pm \text{\tiny{0.497}}\)}}} & \cellcolor{tableblue}{{\textcolor{black}{42.1 \(\pm \text{\tiny{0.233}}\)}}} & {\textcolor{black}{0.153 \(\pm \text{\tiny{0.002}}\)}} & {\textcolor{black}{0.040 \(\pm \text{\tiny{0.001}}\)}} \\
\method (Ours) & \cellcolor{tableblue}{\textbf{\textcolor{black}{47.5 \(\pm \text{\tiny{0.261}}\)}}} & \cellcolor{tableblue}{\textbf{\textcolor{black}{41.0 \(\pm \text{\tiny{0.227}}\)}}} & \textbf{\textcolor{black}{0.145 \(\pm \text{\tiny{0.001}}\)}} & \textbf{\textcolor{black}{0.037 \(\pm \text{\tiny{0.001}}\)}} \\

\bottomrule
\end{tabular}
\endgroup
}
\vspace{-13pt}
\end{table}

\begin{figure}[tbp]
    \centering
    \includegraphics[width=0.92\columnwidth]{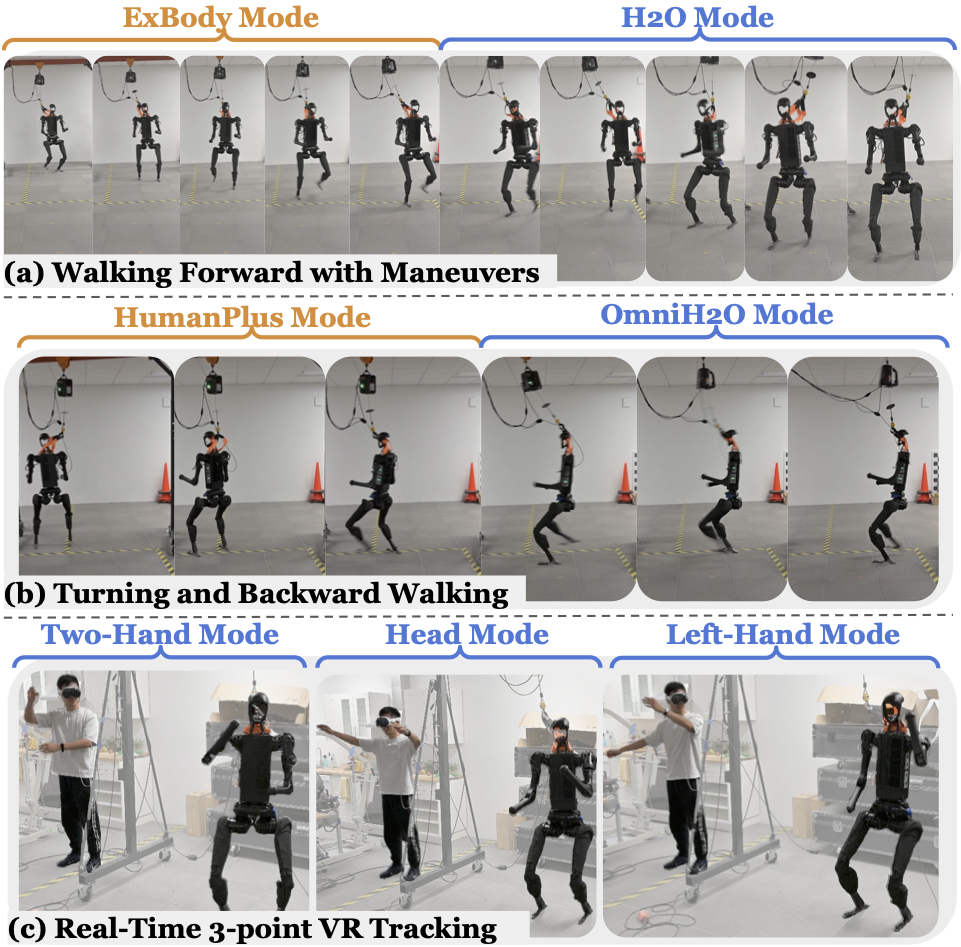}
    \caption{\method shows robustness under control mode switches during locomotion and real-time teleoperation tests.
    }
    \label{fig:locomotion}
    \vspace{-17pt}
\end{figure}

\paragraph{Multi-Mode Evaluations} We also evaluate \method's generalization to locomotion in~\Cref{fig:locomotion}, where we abruptly switch command modes during operation to simulate real-life scenarios. \method successfully transitions from ExBody mode to H2O mode during forward walking in~\Cref{fig:locomotion}(a), and from HumanPlus mode to OmniH2O mode while performing turning and backward walking, in~\Cref{fig:locomotion}(b). Additionally, we conduct a real-test teleoperation demo with Vision Pro, randomly masking out the positions of the head and hands. For example, in the middle of~\Cref{fig:locomotion}(c), the humanoid tracks only the human's head position, ignoring the waving hands in \textit{head mode}. The results demonstrate that \method can smoothly track motions across different modes, showcasing its robustness for real-world scenarios (e.g., when there are occlusions in the reference motions).

\section{RELATED WORK}


\paragraph{Humanoid Whole-Body Controller} Performing whole-body control on humanoid hardware is a long-standing challenge in robotics due to the complex structure of humanoid robots. Before the rise in popularity of learning-based controllers, classical humanoid controllers~\cite{brooks1986robust,gouaillier2009mechatronic, radford2015valkyrie,sakagami2002intelligent, sentis2006whole, ishiguro2020bilateral,dafarra2024icub3,chignoli2021humanoid,okada2005humanoid,kuindersma2016optimization} often use a hierarchical model-based optimization to solve for the low-level torque or position commands sent to hardware motors, where actuator-level dynamics on single joints are abstracted to multi-joint \cite{radford2015valkyrie} or whole-body \cite{sakagami2002intelligent} controllers.
Learning-based controllers follow the same design pattern in spirit, where high-level inputs are translated into low-level motor commands via neural networks. The design of controller abstraction and task specification varies by user needs and applications \cite{darvish2023teleoperation,hauser2024analysis,wonsick2021human}. Recent works on learning-based humanoid whole-body control~\cite{cheng2024expressive,he2024learning,he2024omnih2o,fu2024humanplus,serifi2024vmp} generally have three design patterns for humanoid whole-body controller: kinematic motion tracking \cite{he2024omnih2o, he2024learning,serifi2024vmp}, local joint angle tracking \cite{cheng2024open,fu2024humanplus}, and root velocity tracking \cite{zhuang2024humanoid,li2024reinforcement}.
Kinematic motion tracking means tracking the full-body kinematic motion for each rigid body of the humanoid, and is heavily inspired by the motion imitation in the graphics community \cite{Luo2023-ft, Won2020-lb, Chentanez2018-cw, Peng2018-fu}. Local joint angle tracking tracks the local joint angles of the humanoid, which can be considered a specialized case of kinematic motion imitation where global position information is discarded. Root velocity tracking only serves for locomotion ability, and is used for navigation and terrain traversal of humanoids \cite{zhuang2024humanoid,li2024reinforcement}. One can also combine different control modes for upper and lower body: for instance, the upper body can be controlled with local joint angle tracking and the lower body with velocity tracking \cite{cheng2024expressive}. 
Even within the same kinematic tracking pattern, the sparsity design varies depending on the selected keypoints~\cite{he2024omnih2o,he2024learning}.
So far, each of these control modes is independently developed and are not compatible with each other. In this work, we aim to unify all of these control modes.





\paragraph{Unified Neural Whole-Body Controller for Humanoid}
MHC~\cite{dugar2024learning,shrestha2024generating} learns multi-mode humanoid controller using RL from retargeted motion, but does not support arbitrary subset of chosen modes and is limited to local joint angles and root tracking. 
In computer graphics, MaskedMimic~\cite{tessler2024masked} enables multi-mode control with flexible kinematic tracking constraint by distillation. Other graphics works leverage reusable motion latent space for downstream flexible control modes~\cite{Peng2022-vr, tessler2023calm, Luo2023-er,}. However, additional policies need to be trained.  
In this work, we aim to learn a unified control policy that can be directly used to control real humanoids using different control modes. 

\section{CONCLUSIONS}

In this work, we introduced \method, a unified neural controller for humanoid whole-body control that supports diverse control modes. Through the use of a kinematic motion imitator and policy distillation, \method consolidates motor skills across multiple control modes into a unified policy that outperforms specialized controllers. 
Our evaluations collectively illustrate \method's ability to handle diverse real-world control modes, offering and superior performance compared to specialist policies. 
Future work will explore further developing an automated mode-switching module for real-world applications.




%


\clearpage

\section*{Acknowledgement}
We appreciate Unitree Robotics and Fourier Intelligence for supporting hardware experiments and thank Xue Bin Peng, Chen Tessler, Scott Reed, Viktor Makoviychuk, Yuqi Xie, Avnish Narayan, Zu Wang, Xuxin Cheng, Chong Zhang, Zixuan Chen, and Ziqiao Ma for the insightful discussions.
\printbibliography

\end{document}